\documentclass{article}
\usepackage{ijcai24}

\usepackage[a-3b]{pdfx}
\usepackage{times}
\usepackage{soul}
\usepackage{url}

\usepackage[utf8]{inputenc}
\usepackage[small]{caption}
\usepackage{graphicx}
\usepackage{amsmath}
\usepackage{amsthm}
\usepackage{bm}
\usepackage{booktabs}
\usepackage{algorithm}
\usepackage{algorithmic}
\usepackage[switch]{lineno}
\usepackage{makecell}
\usepackage{multirow}
\usepackage{multicol}
\usepackage{float}                 
\usepackage{subfig}                
\usepackage{overpic}                
\usepackage{amsfonts}
\hypersetup{colorlinks=true, linkcolor=black, urlcolor=black, citecolor=black, filecolor=black}

\title{CAP: A Context-Aware Neural Predictor for NAS}
\author{
Han Ji\and
Yuqi Feng\And
Yanan Sun\footnote{Corresponding author.} \\
\affiliations
College of Computer Science, Sichuan University\\
\emails
jihan@stu.scu.edu.cn,
feng770623@gmail.com,
ysun@scu.edu.cn
}

\begin{document}

\maketitle

\begin{abstract}
     Neural predictors are effective in boosting the time-consuming performance evaluation stage in neural architecture search (NAS), owing to their direct estimation of unseen architectures. Despite the effectiveness, training a powerful neural predictor with fewer annotated architectures remains a huge challenge. In this paper, we propose a context-aware neural predictor (CAP) which only needs a few annotated architectures for training based on the contextual information from the architectures. Specifically, the input architectures are encoded into graphs and the predictor infers the contextual structure around the nodes inside each graph. Then, enhanced by the proposed context-aware self-supervised task, the pre-trained predictor can obtain expressive and generalizable representations of architectures. Therefore, only a few annotated architectures are sufficient for training. Experimental results in different search spaces demonstrate the superior performance of CAP compared with state-of-the-art neural predictors. In particular, CAP can rank architectures precisely at the budget of only 172 annotated architectures in NAS-Bench-101. Moreover, CAP can help find promising architectures in both NAS-Bench-101 and DARTS search spaces on the CIFAR-10 dataset, serving as a useful navigator for NAS to explore the search space efficiently. Our code is available at:~\href{https://github.com/jihan4431/CAP}{https://github.com/jihan4431/CAP}.
\end{abstract}

\section{Introduction}
Neural architecture search (NAS)~\cite{elsken2019neural} has gained increasing attention for its capability to automatically design well-performing deep neural networks in various real-world applications~\cite{sun2019evolving,ghiasi2019fpn,liu2019auto}. However, NAS often needs to evaluate derived architectures with the help of numerous computational resources, which are often not affordable for most researchers. To tackle this issue, various methods are proposed to reduce the search cost, such as early stopping~\cite{ru2021speedy}, using proxy dataset~\cite{zela2020surrogate}, weight sharing~\cite{pham2018efficient}, and so on. Although these methods indeed improve the efficiency of NAS, they suffer severely from performance degradation, making the performance of architectures searched cannot be guaranteed in practice~\cite{xie2023efficient}. 

\begin{figure}[t]
	\centering
        \includegraphics[width=\columnwidth]{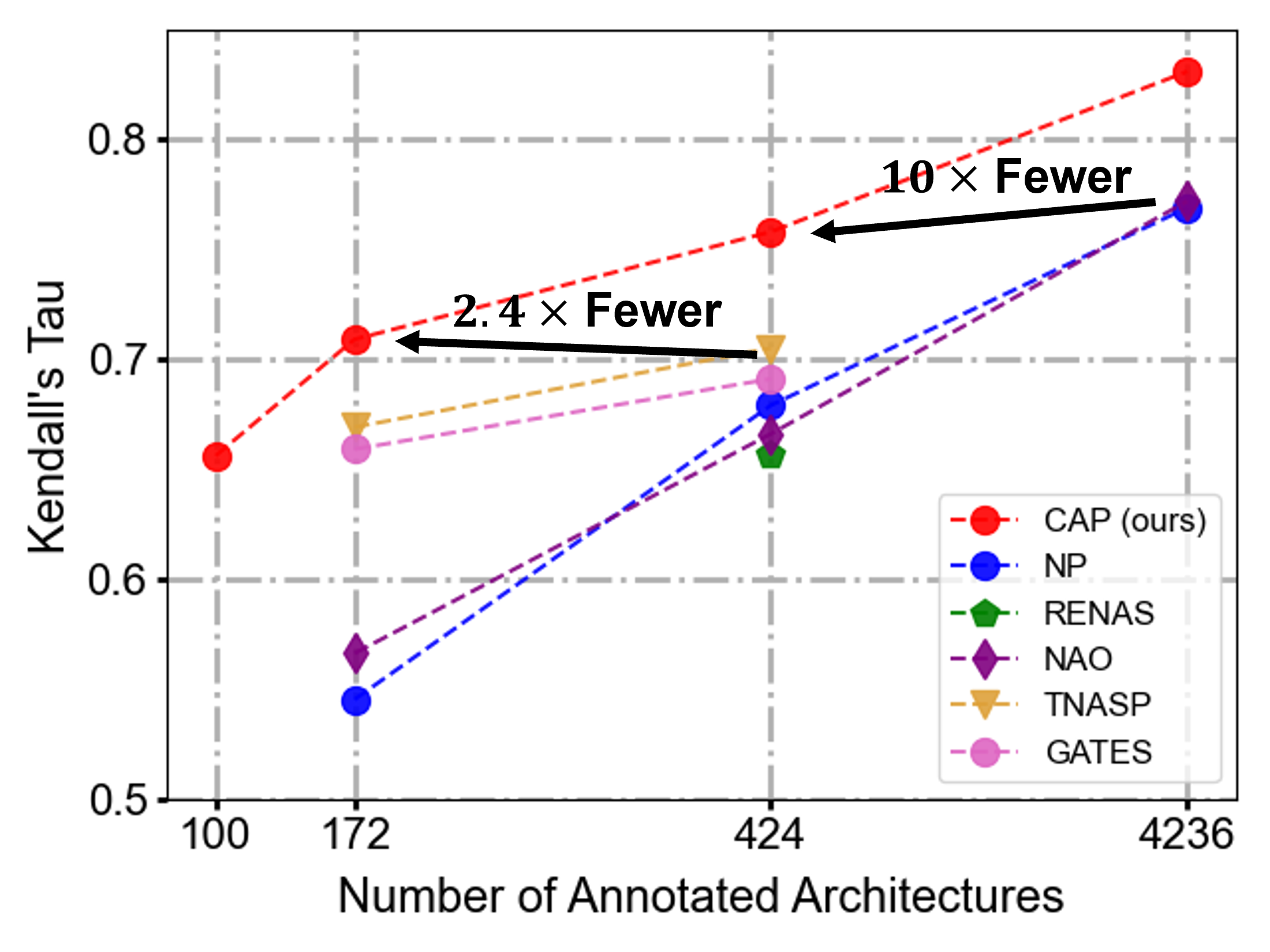}
	\caption{The number of annotated architectures for evaluating all the architectures in NAS-Bench-101 is illustrated. Trained by only 172 annotated architectures, CAP beats most existing predictors which use 424 annotated ones for training. Moreover, CAP utilizes $10\times$ fewer annotated architectures with marginal performance drop compared to other predictors.} 
        \label{fig:Figure 1}
\end{figure}

Neural predictor is a promising acceleration method which can directly estimate the performance of an unseen architecture accurately~\cite{wen2020neural}. In practice, it requires extensive architecture-performance pairs to learn such mapping on the specific tasks. However, these neural predictors tend to perform well only when enough reliable annotated architectures are available, but obtaining such architectures requires a repetitive training process, which is still time-consuming. 

Several works have tried to train a well-performing predictor when a small amount of annotated architectures are available. For example, Semi-NAS~\cite{luo2020semi} attempts to generate more annotated architectures in a semi-supervised way and HAAP~\cite{liu2021homogeneous} proposes a homogeneous architecture augmentation strategy for predictors to expand the training data. However, neither focuses on improving architecture representation to enhance the predictor. We observe that the massive unlabeled architectures in the search space can potentially help the predictor produce expressive representations of architectures but are rarely explored at present. If these data are leveraged properly for training the predictor, it is unnecessary to annotate a large amount of architectures. Namely, the search cost for predictor-based NAS can be greatly reduced. 

In view of the observation above, we propose a context-aware neural predictor (CAP) and design a context-aware self-supervised task for neural predictors to make full use of the substantial unlabeled architectures. By predicting the contextual structure around subgraphs in each architecture, the predictor is enabled to understand and capture its intrinsic topology information and node features. Specifically, we extract the central subgraph and the corresponding context graphs in every graph-like architecture, to construct graph pairs. During this pre-training stage, the graph data from the same pair can be embedded into similar representations. Following the commonly used fine-tuning process in self-supervised learning~\cite{liu2021self}, we apply a simple regression model to map the architecture representation to the real performance and then utilize a few annotated architectures to fine-tune the predictor. Consequently, CAP can capture the topological information at the structure level and the node features at the attribute level of architectures. 

Our contributions can be summarized as follows:
\begin{itemize}
    \item We propose a context-aware neural predictor (CAP) which requires much fewer annotated architectures for training. Compared with the existing neural predictors, CAP can still rank architectures in the search space precisely even with fewer annotated architectures. 
    
    \item  We design a context-aware self-supervised task to pre-train the neural predictor, which is capable of generating meaningful and generalizable representations of architectures without involving annotated architectures. This helps predict the performance of architectures precisely and search for well-preforming architectures in NAS.

    \item The experimental results show that CAP can evaluate the performance ranking of architectures precisely in NAS-Bench-101 and NAS-Bench-201 search spaces with fewer annotated architectures. Specifically, the number of annotated architectures needed by CAP is $2.4\times$ fewer than that of most existing predictors as shown in Figure~\ref{fig:Figure 1}. Moreover, CAP can achieve satisfactory search results in NAS-Bench-101, NAS-Bench-201, and DARTS search spaces. In addition, the ablation studies demonstrate the effectiveness of the proposed context-aware self-supervised task for neural predictors. 
    
\end{itemize}

\section{Related Works}
\subsection{Neural Predictor}
Neural predictor is an effective method to evaluate the performance of architectures and consequently guides the optimal searching direction in NAS. Generally, most of the existing predictors have three stages: 1) Sufficient architectures are collected and then trained from scratch; 2) a predictor is constructed after utilizing these annotated architectures to learn their mapping to the real performance; 3) the predictor can be applied to estimate unseen architectures rapidly and accurately. In these approaches, only a small portion of labeled architectures are selected to construct the predictor due to the unacceptable computation cost required for training. Several works concentrate on making full use of the limited annotated architectures from the perspective of information flow~\cite{ning2020generic}, operation hierarchy~\cite{chen2021not}, and spatial topology information~\cite{lu2021tnasp}, respectively. In contrary to the aforementioned methods, we dive into the substantial but less explored unlabeled architectures. After pre-training the predictor on these highly accessible data to learn their intrinsic features, very limited annotated architectures are enough for a well-performing predictor. 

We note that some previous works try to utilize unlabeled architectures as well, which is similar to our work. Arch2Vec~\cite{yan2020does} uses the vanilla graph reconstruction pretext task to obtain representations from architectures in an unsupervised manner. Instead of focusing on the vanilla graph reconstruction which simply ignores the distinctiveness of the architectures during encoding, we propose a context-aware pretext task to make full use of the rich contextual information from massive unlabeled architectures. To extract node attributes among architectures, GMAE-NAS~\cite{jing2022graph} constructs a difficult graph reconstruction pretext task by masking a large portion of nodes in the architectures to pre-train the predictors. However, GMAE-NAS only focuses on the attribute of operations in the architectures. In contrast, the CAP method further considers the structural information by introducing a context-aware pretext task, enriching the predictor with more valuable prior knowledge.

\subsection{Context-Aware Self-Supervised Learning} 

\begin{figure*}[t!]
    \centering
        \includegraphics[width=\textwidth]{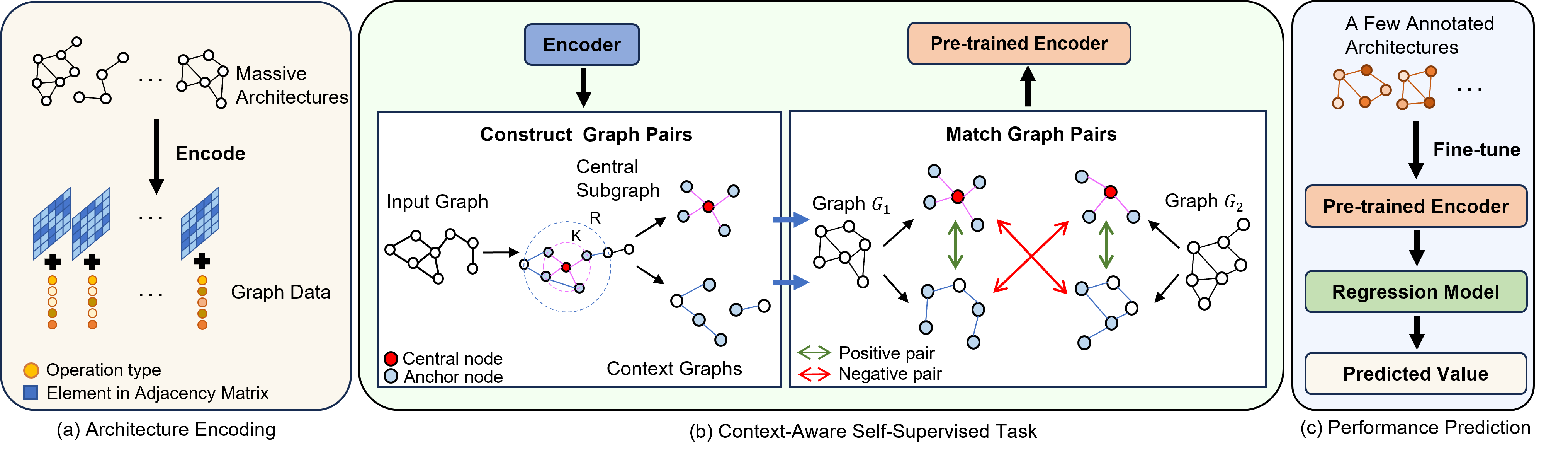}
	
    \caption{Overall framework of the proposed CAP method. (a) First, massive architectures are encoded into graph data. In each architecture, operation types are represented by nodes and their connection ways are denoted by an adjacent matrix. (b) For each input graph data, the central subgraph and corresponding context graphs are extracted to construct a graph pair. Then, the encoder is encouraged to match the graph pairs correctly during the context-aware self-supervised task. For example, the central subgraph and context graphs from $G_1$ are a positive pair. (c) Once the pre-training stage is finished, only a few annotated architectures are used to fine-tune the predictor.}
    \label{fig:Figure 2}
    \end{figure*} 

Self-supervised learning is gaining mounting attention for its successful applications in various domains. Typically, it defines a pretext task based on unlabeled input data to obtain descriptive representation for different downstream tasks. Context-aware self-supervised learning is a significant branch, which constructs pretext tasks mainly according to contextual information inside the data. In natural language processing, a typical context-aware task is to predict the contextual sentences of a given word. In this way, the model is expected to explore the relationship of words in the text without involving any label. Similarly in computer vision, ~\cite{doersch2015unsupervised} predicts the relative positions of contextual patches in an image and enables the model to produce significant feature representation. Moreover, TS-TCC~\cite{eldele2021time} also learns time-series representation from unlabeled data by contextual contrasting. In light of the successful applications above, we propose a context-aware self-supervised task to predict the contextual surrounding in an architecture. The predictor is encouraged to obtain intelligible representations of architectures and thus relies on fewer annotated samples.

\section{Methodology} 
As shown in Figure~\ref{fig:Figure 2}, the overall framework of the proposed CAP method contains three parts, \textit{i.e.}, the architecture encoding, the context-aware self-supervised task, and the performance prediction.

\subsection{Architecture Encoding}
Following the conventions of predictor-based NAS algorithms, a neural architecture can be viewed as a standard graph where the nodes denote the various operation layers and the edges denote the various connection manners. Thus, a $N$-layer architecture $X$ is naturally encoded as an operation list $O_N= \{{o_1,o_2,...,o_N}\}$ and an adjacent matrix $A_{N\times N}$. In the operation list $O_N$, $o_i$ represents the operation type of the $i$-th layer.

However, not all search spaces treat the operations of architectures as nodes in the graph. To bridge the gap, we shift the form of the operation-on-edge search spaces (\textit{e.g.}, the NAS-Bench-201~\cite{dong2019bench} and DARTS~\cite{liu2018darts} search spaces) to the standard version via a simple transformation.

Considering that an architecture could be represented as different isomorphic graphs, we choose a variant version of Graph Isomorphism Networks (GINs)~\cite{xu2018powerful} as the encoder to obtain invariant and expressive representation from given architectures. Generally, the update mechanism of the $k$-th layer of a GIN can be formulated as follows:
\begin{align}   
h_v^{(k)}=&\operatorname{H}_{\theta}\left(\left(1+\epsilon^{(k)}\right) \cdot h_v^{(k-1)}+\sum_{u \in \mathcal{N}(v)} h_u^{(k-1)}\right),
\end{align}%
where $h_v$ and $h_u$ denote the features of the node $v$ and $u$, ${H}_{\theta}$ is a simple MLP, ${N}(v)$ indicates the set of all nodes within the neighborhood of node $v$, and $\epsilon$ is a learnable weight.

\subsection{Context-Aware Self-Supervised Task}
Once the architectures are encoded into graph data, we conduct the context-aware self-supervised task to pre-train the predictor. As for graph self-supervised learning, there exist two kinds of popular pretext tasks: predictive pretext tasks and contrastive pretext tasks. The former aim to predict the self-generated labels to provide self-supervision for the model. The latter produce various augmentation views for a given graph sample and then maximize the similarity of those matching pairs generated from the same sample. We select the latter to leverage the rich contextual information among architectures because contrastive learning methods are on the ground of mutual information maximization and predictive learning methods lack a unified theoretical framework at present~\cite{xie2022self}. Inspired by the application of successful context-aware self-supervised tasks in natural language processing~\cite{mikolov2013distributed}, computer vision~\cite{doersch2015unsupervised} and times-series~\cite{eldele2021time}, we allow the encoder to predict the contextual surrounding of any node in each architecture. The basic idea is to first extract the central subgraph and the context graphs from each architecture to construct graph pairs. After that, the similarity between matching pairs which are obtained from the same architectures is maximized. 

In the context-aware self-supervised task, we randomly choose a central node $v$ inside the graph which denotes an architecture. Then, we exploit its $K$-hop neighborhood $G_v^{K}$ to represent the selected node, containing all the nodes and edges at most $K$ hops away from node $v$. Once given the central node $v$ and its neighborhood $G_v^{K}$, the surrounding context graphs are confirmed accordingly. We introduce a distance scale $R$ ($R>K$) as the higher bound of context graphs. To be specific, all subgraphs between the $K$-hop neighborhood and $R$-hop neighborhood of the central node are regarded as the context graphs. Notably, some nodes are shared between the $K$-hop neighborhood and the outer context graphs, which are called anchor nodes, functioning as a bridge to convey information from the central node $v$ to its context graphs.

For context-aware abilities, the encoders learn to match the neighborhood representation to the corresponding context representation inside an architecture by maximizing their similarity. Therefore, the $K$-hop neighborhood and its outer context graphs are first embedded into $h_v^{G}$ and $c_v^{G}$ by the encoders, respectively. Similar to other context-aware tasks, these context graphs are aggregated to represent the contextual information. Therefore, an auxiliary GIN is necessary to aggregate representation from context graphs of different sizes next to the neighborhood and generate a representative embedding $c_v^{G}$ for the similarity computation.

Following the common practice in contrastive learning, we introduce the negative sampling strategy to train a discriminating GIN encoder. In the proposed context-aware self-supervised task, we expect the encoder to predict the contextual structure based on the central subgraph. Hence, we randomly choose the embeddings of central subgraphs from other graphs as negative samples. We take the ($h_v^{G}$, $c_v^{G'}$) pairs from the same graphs as the positive ones (${G}={G'}$) and those from different graphs as the negative ones (${G}\neq{G'}$). To be specific, the negative sampling ratio is set to one, achieving the balance between positive pairs and negative ones. The loss function of the context-aware self-supervised task can be formulated as:
\begin{align}
    L_{ss} = & L_{CE}(\vec{1},\operatorname{sim}(h_v^{G},c_v^{G}))+L_{CE}(\vec{0},\operatorname{sim}(h_v^{G},c_v^{G'})),
\end{align}%
where $(h_v^{G},c_v^{G})$ is a positive pair from the same graph $G$ while $(h_v^{G},c_v^{G'})$ composes a negative pair; $L_{CE}$ denotes the cross entropy loss function.

After that, a readout function is added in the tail to obtain the graph-level representation of every architecture $X$:
\begin{align}
    h_X^{(k)} = & \operatorname{readout}(\{h_{v_i}^{(k)}|v_i\in X\}),
\end{align}%

The readout function here can be various graph-level pooling operations, we use averaging here for its simplicity. 

The pre-trained main GIN model can be directly used as the encoder for the neural predictor. Additionally, the auxiliary GIN model is only used in the pre-training procedure and discarded in the following performance prediction.

\subsection{Performance Prediction}
After the pre-training process, a regression model is required to evaluate the real performance based on the representations of corresponding architectures. For simplicity, we choose a two-layer MLP as the regression model.

To fine-tune the pre-trained predictor with limited annotated architectures, we select decoder-only fine-tuning, full fine-tuning, and partial fine-tuning methods. 

\paragraph{Decoder-only Fine-tuning.} All weights of parameters in the encoder are fixed once pre-training is done. During the fine-tuning, only the regression model is trained while the encoder is frozen. In other words, the representation of any given architecture remains unchanged. 

\paragraph{Full Fine-tuning.} The whole pre-trained encoder and the regression model are jointly trained with a few labeled architectures. In this way, the pre-training task helps to initialize the encoder with better weights in comparison to random initialization.

\paragraph{Partial Fine-tuning.} During the fine-tuning stage, partial fine-tuning discards the dropout layers and batch normalization layers in the pre-trained encoder. Thus, it takes less time to fine-tune the predictor compared with the full fine-tuning method.  

As for the loss function, mean squared error (MSE) and ranking loss are two popular choices in previous works of neural predictors. Intuitively, MSE between the absolute performance of an architecture and the predicted value is effective. In this way, the model is expected to predict the real performance precisely. Unfortunately, a low MSE loss may mislead the predictors. MSE loss is unable to reflect the correct ranking between the queried architectures when their accuracy is close~\cite{xu2021renas}. Conversely, it is more important to focus on the relative performance ranking of architectures considering that the goal is to search for the best architecture instead of its absolute performance in the NAS methods. Here, we use Bayesian personalized ranking (BPR)~\cite{rendle2012bpr} loss for the sake of its popularity in relative ranking tasks. The loss function can be formulated as follows:
\begin{align}
    L_{pp} = & \sum_{(A_i, A_j) \in D} \log \sigma\left(s(A_i)-s(A_j)\right),
\end{align}%
where $D$ represents all the architecture pairs like $(A_i,A_j)$ in the training set, ${s(A_i)}$ is the predicted performance score of the architecture $A_i$, and $\sigma$ denotes the sigmoid function.

\section{Experiments} 
The experiments performed contain three parts in terms of the ranking capability, the search results, and the ablation studies. In the first part, we verify the powerful ranking capability of CAP on NAS-Bench-101~\cite{ying2019bench} and NAS-Bench-201~\cite{dong2019bench} benchmarks, especially when the training samples are extremely limited. In the second part, we use both closed domain search spaces (NAS-Bench-101, NAS-Bench-201) and open domain search space (DARTS~\cite{liu2018darts}) to search for promising architectures on CIFAR-10, CIFAR-100~\cite{Krizhevsky2009LearningML}, and ImageNet16-120~\cite{chrabaszcz2017downsampled}. In the third part, we further investigate the effectiveness of the proposed context-aware self-supervised task via detailed ablation studies. 

\subsection{Experimental Settings} 
\paragraph{NAS-Bench-101 Search Space.} The NAS-Bench-101 is a cell-based search space consisting of over 423k unique neural architectures. In each architecture cell, at most seven nodes and nine edges are contained. The nodes represent operation layers and the edges represent connection manners. The search space provides the validation accuracy and test accuracy of convolutional neural networks (CNNs) on CIFAR-10. 

\paragraph{NAS-Bench-201 Search Space.} The NAS-Bench-201 is also a cell-based search space constructed by over 15K different neural architectures. Each cell includes four nodes and six edges. Different from the operation-on-node NAS-Bench-101 search space, the operations are represented by edges in NAS-Bench-201. Detailed results of CNNs are available on three datasets: CIFAR-10, CIFAR-100, and ImageNet16-120.

\paragraph{DARTS Search Space.} The DARTS search space is built by both normal cells and reduction cells where each cell is composed of seven nodes and 14 edges. It is a much larger open domain search space than the NAS-Bench-101 and the NAS-Bench-201 search spaces. In the DARTS search space, the operations are also represented by edges.

\subsection{Ranking Results on NAS Benchmarks}
The experiments are carried out on both NAS-Bench-101 and NAS-Bench-201 benchmarks. In parallel to previous works~\cite{liu2021homogeneous,lu2021tnasp}, this work utilizes the validation accuracy of an architecture to train the predictor and evaluates its corresponding test accuracy. In terms of metrics for ranking ability, we choose Kendall’s Tau between the predicted accuracy of architectures and the real accuracy for its widespread use in prior works. For fair comparisons among different neural predictors, we adopt the same data splits in TNASP~\cite{lu2021tnasp} for both NAS-Bench-101 and NAS-Bench-201.

\begin{table}[h]
    \centering
    \resizebox{\columnwidth}{!}{
    \begin{tabular}{cccccc}
        \toprule
        \textbf{Train Samples}   & 100 & 172 & 424 & 424 & 4236 \\
        \textbf{Validation Samples}  & 200  & 200  & 200  & 200  & 200 \\
        \textbf{Test Samples}    & all  & all & 100 & all & all \\    
        \midrule
        ReNAS$^{\dagger}$~\cite{xu2021renas}            & -     & -     & 0.634 & 0.657 & 0.816        \\  
        NP$^{\ddagger}$ ~\cite{wen2020neural}              & 0.391 & 0.545 & 0.710 & 0.679 & 0.769        \\
        NAO$^{\ddagger}$~\cite{luo2018neural}              & 0.501 & 0.566 & 0.704 & 0.666 & 0.775        \\
        Arch2Vec$^{\star}$~\cite{yan2020does}         & 0.435 & 0.511 & 0.561 & 0.547 & 0.596        \\
        GATES$^{\star}$~\cite{ning2020generic}            & 0.605 & 0.659 & 0.666 & 0.691 & 0.822        \\
        CTNAS \cite{chen2021contrastive}            & -     & -     & 0.751 & -     & -            \\
        TNASP \cite{lu2021tnasp}         & 0.600 & 0.669 & 0.752 & 0.705 & 0.820        \\
        TNASP-SE \cite{lu2021tnasp}         & 0.613 & 0.671 & 0.754 & 0.722 & 0.820        \\
        GMAE-NAS$^{\star}$~\cite{jing2022graph}         & 0.626 & 0.687 & 0.778 & 0.733 & 0.775        \\
        \midrule
        \textbf{CAP}            & \textbf{0.656} & \textbf{0.709} & \textbf{0.791} & \textbf{0.758} & \textbf{0.831}        \\
        \bottomrule
    \end{tabular}
    }
    \caption{Ranking results of different data splits on NAS-Bench-101 benchmark. The Kendall’s Tau of 10 independent runs is calculated. $^{\dagger}$: reported by CTNAS. $^{\ddagger}$: reported by TNASP.  $^{\star}$: implemented by ourselves using open-source code.}
    \label{tab:Table 1}
\end{table}

\begin{table}[h]
    \centering
    \resizebox{\columnwidth}{!}{
    \begin{tabular}{cccccc}
        \toprule
        \textbf{Train Samples}   & 78 & 156 & 469 & 781 & 1563 \\
        \textbf{Validation Samples} & 200  & 200  & 200  & 200  & 200 \\
        \textbf{Test Samples}  & all  & all & all & all & all \\    
        \midrule
        NP$^{\dagger}$ ~\cite{wen2020neural}               & 0.343 & 0.413 & 0.584 & 0.634 & 0.646        \\
        NAO$^{\dagger}$~\cite{luo2018neural}              & 0.467 & 0.493 & 0.470 & 0.522 & 0.526        \\  
        Arch2Vec$^{\star}$~\cite{yan2020does}         & 0.542 & 0.573 & 0.601 & 0.606 & 0.605        \\
        TNASP \cite{lu2021tnasp}        & 0.539 & 0.589 & 0.640 & 0.689 & 0.724        \\
        TNASP-SE \cite{lu2021tnasp}        & 0.565 & 0.594 & 0.642 & 0.690 & 0.726        \\
        \midrule
        \textbf{CAP}   & \textbf{0.600} & \textbf{0.684} & \textbf{0.755} & \textbf{0.776} & \textbf{0.815}        \\
        \bottomrule
    \end{tabular}
    }
    \caption{Ranking results of different data splits on NAS-Bench-201 benchmark. The Kendall’s Tau of 10 independent runs is calculated. $^{\dagger}$: reported by TNASP.  $^{\star}$: implemented by ourselves using open-source code.}
    \label{tab:Table 2}
\end{table}

The ranking results on NAS-Bench-101 are shown in Table~\ref{tab:Table 1}. It can be seen that CAP outperforms the other predictors across various data splits. Notably, with only 172 annotated architectures, CAP is on par with the competitors which are trained by 424 training samples and can even beat most of them. Such promising results imply that the contextual information from unlabeled architectures is leveraged in the pre-training stage. Therefore, the pre-trained encoder can produce meaningful representations of architectures even before the fine-tuning stage. As a result, fine-tuning the predictor is much easier compared with training it from scratch. With the pre-trained predictor, only a few annotated architectures are enough to construct a well-performed neural predictor. The satisfactory ranking results demonstrate the effectiveness of the proposed context-aware self-supervised task.

Moreover, we report the ranking results on NAS-Bench-201 in Table~\ref{tab:Table 2}. It can be observed that CAP achieves state-of-the-art performance when compared with other competitors regardless of the data split. It is noteworthy that CAP surpasses them by a large margin, especially in the condition of limited training samples, which is consistent with our previous analysis of the effect of the context-aware self-supervised task. Meanwhile, the proposed CAP can achieve a Kendall’s Tau value of 0.684 when trained by only 156 annotated architectures, which is superior to other predictors trained by 469 ($2.4\times$ more than 56) training samples. The experimental results show that the proposed context-aware self-supervised task helps construct a powerful neural predictor with fewer annotated architectures. We observe that as the number of training samples increases, the gap between CAP and other predictors is gradually narrowed. However, we mainly focus on the performance of the predictor in the few-shot scenario because annotating architectures is extremely time-consuming.

\subsection{Searching on Closed Domain Search Spaces}
We combine the proposed CAP with the regularized evolutionary algorithm~\cite{real2019regularized} to search for promising CNNs in the NAS-Bench-101 benchmark. The total number of queried architectures is set to 150 and the search results are shown in Table~\ref{tab:Table 3}. The traditional methods are placed in the first row, and the second row shows the search results of previous state-of-the-art methods. Compared with them, CAP helps search for better architectures with the same queried budget. Through 10 independent runs, we also find that the CNNs searched by the CAP method achieve $94.18\%$ average test accuracy on CIFAR-10, which is $0.38\%$ higher than the result of the original regularized evolutionary algorithm. These promising experimental results demonstrate that the pre-trained predictor, which serves as the optimal guidance, facilitates searching for promising architectures.

\begin{table}[hb]
    \centering
    \resizebox{\columnwidth}{!}{
    \begin{tabular}{ccc}
        \toprule
        \textbf{Method}    & \textbf{Test Acc.(\%)} & \textbf{Searching Strategy} \\
        \midrule
        RS~\cite{bergstra2012random} & 93.64      &Random                    \\
        REA~\cite{real2019regularized}           & 93.80      &Evolution                \\  
        \midrule
        BANANAS~\cite{white2021bananas}       & 94.08     &Bayesian Optimization                   \\     
        CATE~\cite{yan2021cate}          & 94.12     &REINFORCE                  \\  
        GMAE-NAS~\cite{jing2022graph}      & 94.14     &Evolution              \\
        BONAS~\cite{shi2020bridging}       & 94.14     &Bayesian Optimization  \\
        NPENAS~\cite{wei2022npenas}        & 94.15     &Evolution                \\
        \midrule
        \textbf{CAP}  & \textbf{94.18}   &Evolution                     \\
        \bottomrule
    \end{tabular}
    }
    \caption{Comparison with other NAS methods on CIFAR-10 using the NAS-Bench-101 search space. The number of queried architectures is set to 150. It reports the average test accuracy of 10 independent runs.}
    \label{tab:Table 3}
\end{table}

\begin{table*}[t!]
    \centering
    \resizebox{\textwidth}{!}{
    \begin{tabular}{cccccccc}
        \toprule
        \rule{0pt}{10pt}
        \multirow{2}{*}{\textbf{Method}}   & \textbf{Search Cost} & \multicolumn{2}{c}{\textbf{CIFAR-10}}  & \multicolumn{2}{c}{\textbf{CIFAR-100}} 
        & \multicolumn{2}{c}{\textbf{ImageNet16-120}}  \\
        \cline{3-8}
        \rule{0pt}{10pt}
                                  &\textbf{(Seconds)} &\textbf{Valid} &\textbf{Test} &\textbf{Valid} &\textbf{Test} &\textbf{Valid} &\textbf{Test} \\
     
        \midrule
        ResNet~\cite{he2016deep}        & - & 90.83 & 93.97 & 70.42 & 70.86 & 44.53  & 43.63                    \\
        \midrule
        RS~\cite{bergstra2012random}           & 22993.93 &  $90.93\pm0.36$ & $93.80\pm0.36$ & $70.93\pm1.09$ & $71.04\pm1.07$ & $44.45\pm1.10$  & $44.57\pm1.25$                   \\  
        RL~\cite{williams1992simple}       & 27870.7  & $91.09\pm0.37$ & $93.85\pm0.37$ & $71.61\pm1.12$ & $71.71\pm1.09$ & $45.05\pm1.02$ & $45.24\pm1.18$                   \\
        \midrule
        ENAS~\cite{pham2018efficient}       & 13314.51  & $37.51\pm3.19$ & $53.89\pm0.58$ & $13.37\pm2.35$ & $13.96\pm2.33$ & $15.06\pm1.95$ & $14.84\pm2.10$                   \\
        DARTS-V1~\cite{liu2018darts}       & 10889.87  & $39.77\pm0.00$ & $54.30\pm0.00$ & $15.03\pm0.00$ & $15.61\pm0.00$ & $16.43\pm0.00$ & $16.32\pm0.00$                   \\
        DARTS-V2~\cite{liu2018darts}       & 29901.67  & $39.77\pm0.00$ & $54.30\pm0.00$ & $15.03\pm0.00$ & $15.61\pm0.00$ & $16.43\pm0.00$ & $16.32\pm0.00$                   \\
        GDAS~\cite{dong2019searching}       & 31609.80  & $89.89\pm0.08$ & $93.61\pm0.09$ & $71.34\pm0.04$ & $70.70\pm0.30$ & $41.59\pm1.33$ & 
        $41.71\pm0.98$                   \\
        Arch2Vec-BO~\cite{yan2020does}       & 12000.00  & $91.41\pm0.22$ & $94.18\pm0.24$ & $73.35\pm0.32$ & $73.37\pm0.30$ & $46.34\pm0.18$ & $46.27\pm0.37$ \\
        FairNAS~\cite{chu2021fairnas}       & 9845.00  & $90.97\pm0.57$ & $93.23\pm0.18$ & $70.94\pm0.94$ & $71.00\pm1.46$ & $41.09\pm1.00$ & $42.19\pm0.31$                   \\
        RMI-NAS~\cite{zheng2022neural}       & 1258.21 & $91.44\pm0.09$ & $94.28\pm0.10$ & $73.38\pm0.14$ & $\textbf73.36\pm0.19$ & $46.37\pm0.00$ & $46.34\pm0.00$                   \\
        \midrule
        {Jacob{\_}cov}~\cite{mellor2021neural}       & -  & $89.69\pm0.73$ & $92.96\pm0.80$ & $69.87\pm1.22$ & $70.03\pm1.16$ & $43.99\pm2.05$ & $44.43\pm2.07$                   \\
        Mag~\cite{tanaka2020pruning}       & -  & $89.94\pm0.34$ & $93.35\pm0.04$ & $70.18\pm0.66$ & $70.47\pm0.19$ & $42.57\pm2.14$ & $43.24\pm1.18$                   \\
        \midrule
        \textbf{CAP (Average)}           &\textbf{15.65}  & $\textbf{91.54}\pm\textbf{0.10}$  & $\textbf{94.34}\pm\textbf{0.06}$ & $\textbf{73.41}\pm\textbf{0.17}$ & $\textbf{73.41}\pm\textbf{0.22}$ & $\textbf{46.47}\pm\textbf{0.07}$ & $\textbf{46.44}\pm\textbf{0.36}$                   \\
        CAP (Best)         & - & 91.61 & 94.37 & 73.49 & 73.51 & 46.56  & 46.73            \\
        \midrule
        Optimal & - & 91.61 & 94.37 & 73.49 & 73.51 & 46.73  & 47.31 \\
        \bottomrule
        
    \end{tabular}
    }
    \caption{Comparison with other NAS methods on CIFAR-10, CIFAR-100, and ImageNet16-120 using the NAS-Bench-201 search space. The average searching cost and classification accuracy of the top-1 searched architecture are reported.}
    \label{tab:Table 4}
\end{table*}

On NAS-Bench-201, we follow the settings in~\cite{dong2019bench} for a fair comparison with the baselines. Specifically, 50 annotated architectures are randomly chosen to train the predictor, and the predictor afterwards evaluates all the architectures in the search space. The best performance of the top 50 evaluated architectures is reported as the search result. As demonstrated in Table~\ref{tab:Table 4}, our method can complete the searching within 16 seconds, consuming over 80$\times$ less than the time of baselines. At the same time, the best architecture we search for exceeds the competitors in terms of accuracy. These extraordinary results can be attributed to the sufficient prior knowledge that the predictor learns in advance through the context-aware self-supervised task. Meanwhile, the applied searching strategy leverages the ranking capability of the predictor and thus dramatically cuts the searching cost.

The evaluation strategies during the search can be roughly divided into conventional training-based strategies like reinforcement learning and training-free strategies like zero-cost proxies. It can be observed from Table~\ref{tab:Table 4} that the former tend to discover architectures with stable performance on different independent runs at the price of enormous searching costs. While the latter are sensitive to varied settings and therefore result in a huge discrepancy of search results across the runs. Compared With them, the CAP method achieves the most balanced results on CIFAR-10, CIFAR-100, and ImageNet16-120 and the searched architectures achieve state-of-the-art performance, further verifying the positive effects of the proposed context-aware self-supervised task. 

\begin{table}[h]
    \centering
    \resizebox{\columnwidth}{!}{
    \begin{tabular}{ccccc}
        \toprule
        \textbf{Method}  & \textbf{P(M)} & \textbf{Best Acc.(\%)} & \textbf{Average Acc.(\%)} & \textbf{Cost} \\  
        \midrule
        NASNet-A~\cite{zoph2018learning}         & \textbf{3.3} & 97.35 & - & 1800        \\
        ENAS~\cite{pham2018efficient}             & 4.6 & 97.11 & - & 0.5        \\
        DARTS~\cite{liu2018darts}            & \textbf{3.3} & - & $97.24\pm0.09$ & 4       \\
        \midrule
        NAO~\cite{luo2018neural}         & 10.6 & 97.50 & - & 200       \\
        GATES~\cite{ning2020generic}         & 4.1 & 97.42 & - & -        \\
        Arch2Vec-BO~\cite{yan2020does}   & 3.6 & 97.52 & $97.44\pm0.05$ & 100        \\
        BONAS-D~\cite{shi2020bridging}         & \textbf{3.3} & 97.57 & - & 10.0        \\
        NAS-BOWL~\cite{ru2020interpretable}         & 3.7 & 97.50 & $97.39\pm0.08$ & 3        \\
        BANANAS~\cite{white2021bananas}         & 3.6 & - & $97.33\pm0.07$  & 11.8        \\
        CATE~\cite{white2021bananas}             & 3.5 & - & $97.45\pm0.08$ & 3.3        \\
        CTNAS~\cite{chen2021contrastive} & 3.6 & - & $97.41\pm0.04$ & 0.3        \\
        TNASP~\cite{lu2021tnasp}         & 3.6 & 97.48 & $97.43\pm0.04$ & 0.3        \\
        NPENAS-NP~\cite{wei2022npenas} & 3.5 & 97.56 & $97.46\pm0.10$ & 1.8        \\
        NAR-Former~\cite{yi2023nar}       & 3.8 & 97.52 & - & \textbf{0.24}        \\
        \midrule
        \textbf{CAP}            & \textbf{3.3} & \textbf{97.58} &  \textbf{$\textbf{97.46}\pm\textbf{0.09}$} & 3.3        \\
        \bottomrule              
    \end{tabular}
    }
    \caption{Comparison with other NAS methods on CIFAR-10 using the DARTS search space. Search cost is measured by the GPU Days.}
    \label{tab:Table 5}
\end{table}

\subsection{Searching on Open Domain Search Space}

To search for well-performed CNNs in the DARTS search space, we follow the main procedures in CATE~\cite{yan2021cate}. Despite the massive available architecture performances in the NAS-Bench-301, their accuracy can be inaccurate for they are predicted by a surrogate model trained on 60k architectures. Consequently, we choose to train the architectures from scratch to obtain reliable annotated data for the predictor. Specifically, we randomly choose 100 architectures in the DARTS and then train them from scratch to construct the architecture-performance pairs as training samples. For each candidate architecture, the epoch number is set to 50 with a batch size of 96. As for the searching method, we sample 100k architectures at random in the DARTS search space and they are afterwards evaluated by our trained predictor. After that, we select the architectures with top-3 predicted performance as the search results and re-train them with common DARTS strategies. The search results on CIFAR-10 are reported in Table~\ref{tab:Table 5}.

For a fair comparison, we report the average test accuracy of the top-3 architectures along with the best accuracy. The first row shows the performance of well-known NAS methods, and predictor-based NAS methods are placed in the second row. It can be observed that CAP obtains the best accuracy of $97.58\%$ and the average accuracy of $97.46\%$, outperforming all the others with the least model parameters of 3.3M. More detailed experimental settings for the proposed pretext task and the performance evaluation are reported in the supplementary materials. 

\subsection{Ablation Study}
We conduct ablation studies on the proposed CAP in terms of different hyperparameters using the NAS-Bench-101 search space. $S=\{S1,S2,S3,S4,S5\}$  denotes the data splits in Table\ref{tab:Table 1}. The Kendall’s Tau values of predicted accuracy of architectures are reported. We also compare the searching results of CAP with/without pre-training in the supplementary materials.

\begin{figure}[ht]
    \begin{minipage}[t]{0.5\columnwidth}
        \centering
        \includegraphics[width=\textwidth]{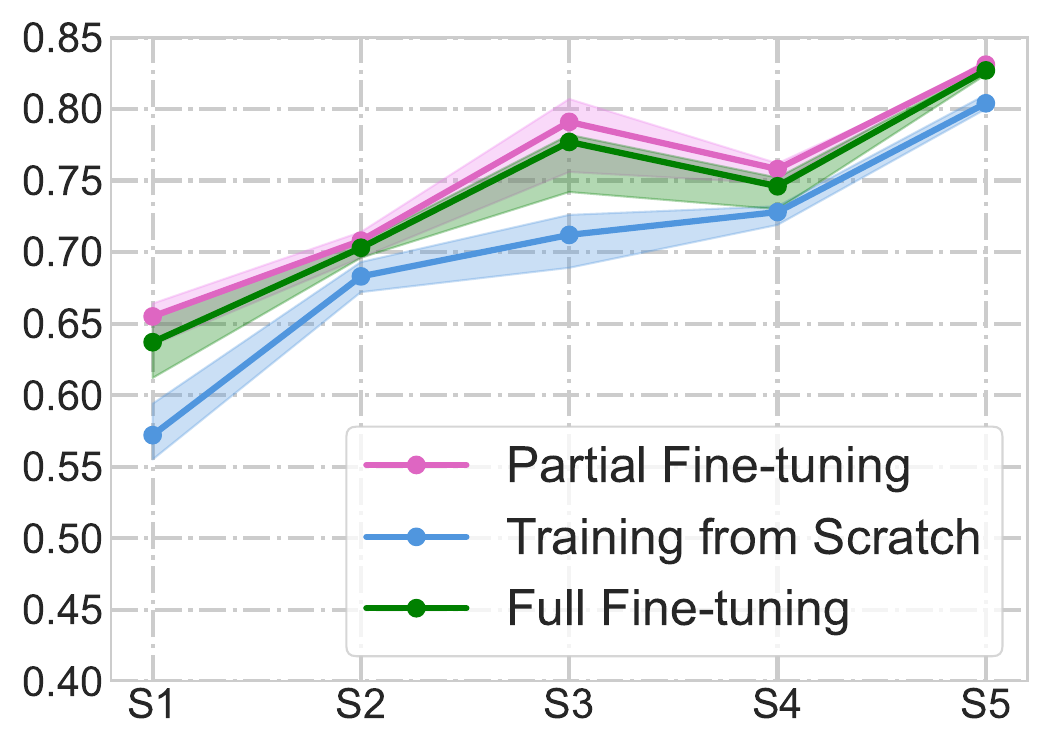}
        \centerline{(a) Fine-tuning Methods}
        \label{fig:Figure_a 4}
    \end{minipage}%
    \begin{minipage}[t]{0.5\columnwidth}
        \centering
        \includegraphics[width=\textwidth]{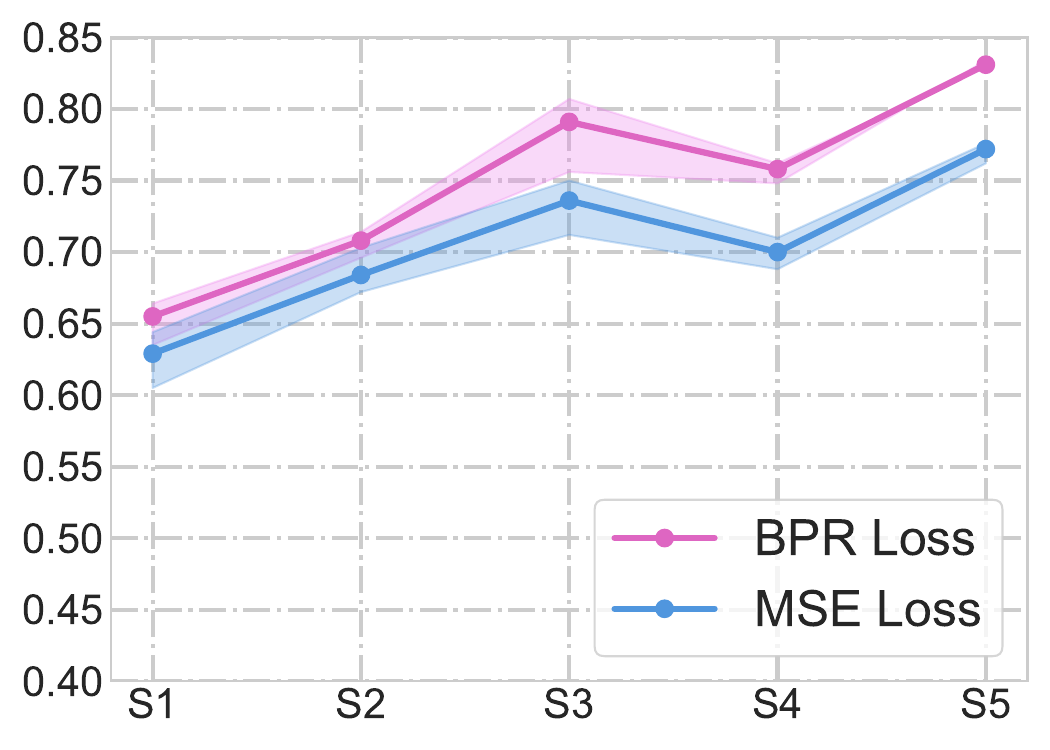}
        \centerline{(b) Loss Functions}
         \label{fig:Figure_b 4}
    \end{minipage}
     \begin{minipage}[t]{0.5\columnwidth}
        \centering
        \includegraphics[width=\textwidth]{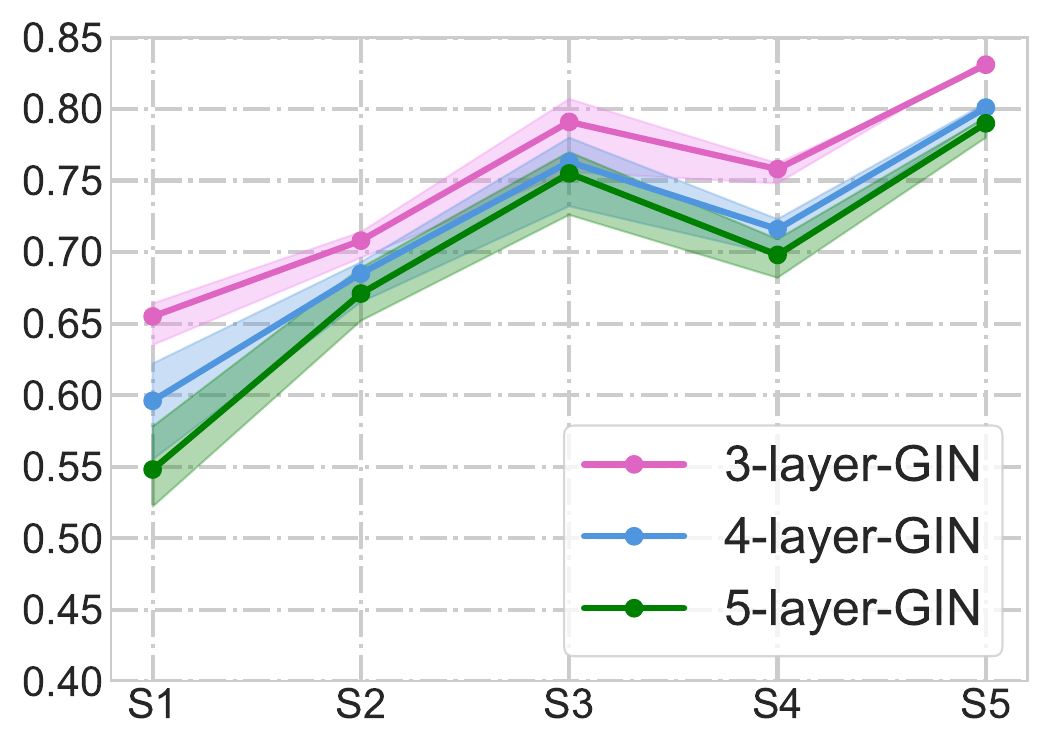}
        \centerline{\makecell{(c) Number of Encoder \\ Layers}}
         \label{fig:Figure_c 4}
    \end{minipage}%
    \begin{minipage}[t]{0.5\columnwidth}
        \centering
        \includegraphics[width=\textwidth]{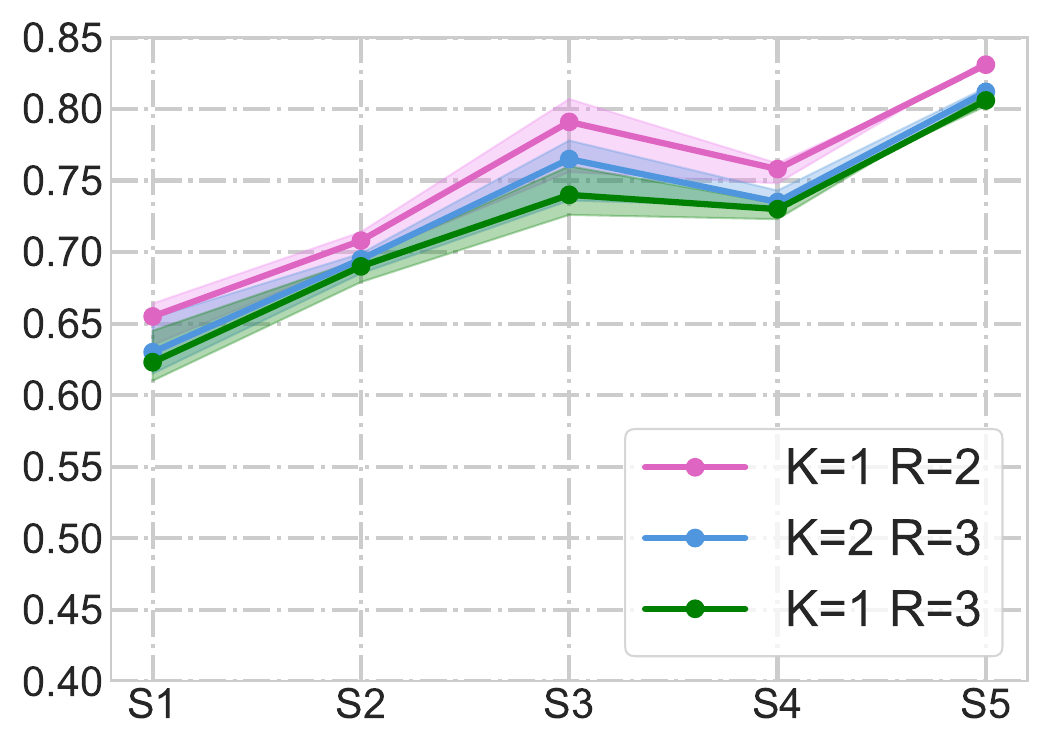}
        \centerline{\makecell{(d) Hyperparameters in \\Pre-training Task}}
         \label{fig:Figure_d 4}
    \end{minipage}
    \caption{Results of Ablation study with different data splits. The Kendall’s Tau of 5 independent runs is calculated.}
    \label{fig:Figure 3}
\end{figure}

\paragraph{Fine-tuning Methods.}
 We train the predictor from scratch to compare it with the pre-trained versions. Furthermore, we explore two different fine-tuning methods for the pre-trained predictor, and the results are shown in Figure~\ref{fig:Figure 3}(a). It's observed that the pre-trained predictor is greatly enhanced when the encoder and the regression model are jointly fine-tuned (\textit{e.g.}, partial fine-tuning and full fine-tuning method). This can be credited to the rich contextual information introduced from architectures through the context-aware self-supervised task. The decoder-only fine-tuning method performs poorly because the encoder keeps frozen and lacks supervision from annotated architectures when fine-tuning, whose results are included in the supplementary materials. We also discover that the partial fine-tuning method achieves better performance than the full fine-tuning method because of a faster convergence rate. 

\paragraph{Loss Functions.}
We adopt MSE loss and BPR loss for training the predictor in the fine-tuning stage. According to Figure~\ref{fig:Figure 3}(b), the predictor trained by BPR loss achieves better ranking results across various training data splits. We attribute this to its emphasis on the relative ranking of architectures rather than their absolute performance valves. 

\paragraph{Number of Encoder Layers.}
We investigate the number of layers $L$ in the GIN encoder for the predictor and report the ranking results in Figure~\ref{fig:Figure 3}(c). The optimal number $L$ is three and deeper GIN leads to the decrease in predictor performance.    

\paragraph{Hyperparameters in Pre-Training Task.}
 In the proposed context-aware self-supervised task, the choice of $K$ and $R$ also has effects on the ranking results of the predictor. They are summarized in Figure~\ref{fig:Figure 3}(d). We find that the ranking performance is satisfactory when the central subgraph in each architecture and the corresponding context graphs share only a few overlapping nodes (\emph{i.e.} $K=1, R=2$).

\begin{figure}[htbp]
    \centering
    \includegraphics[width=0.95\columnwidth]{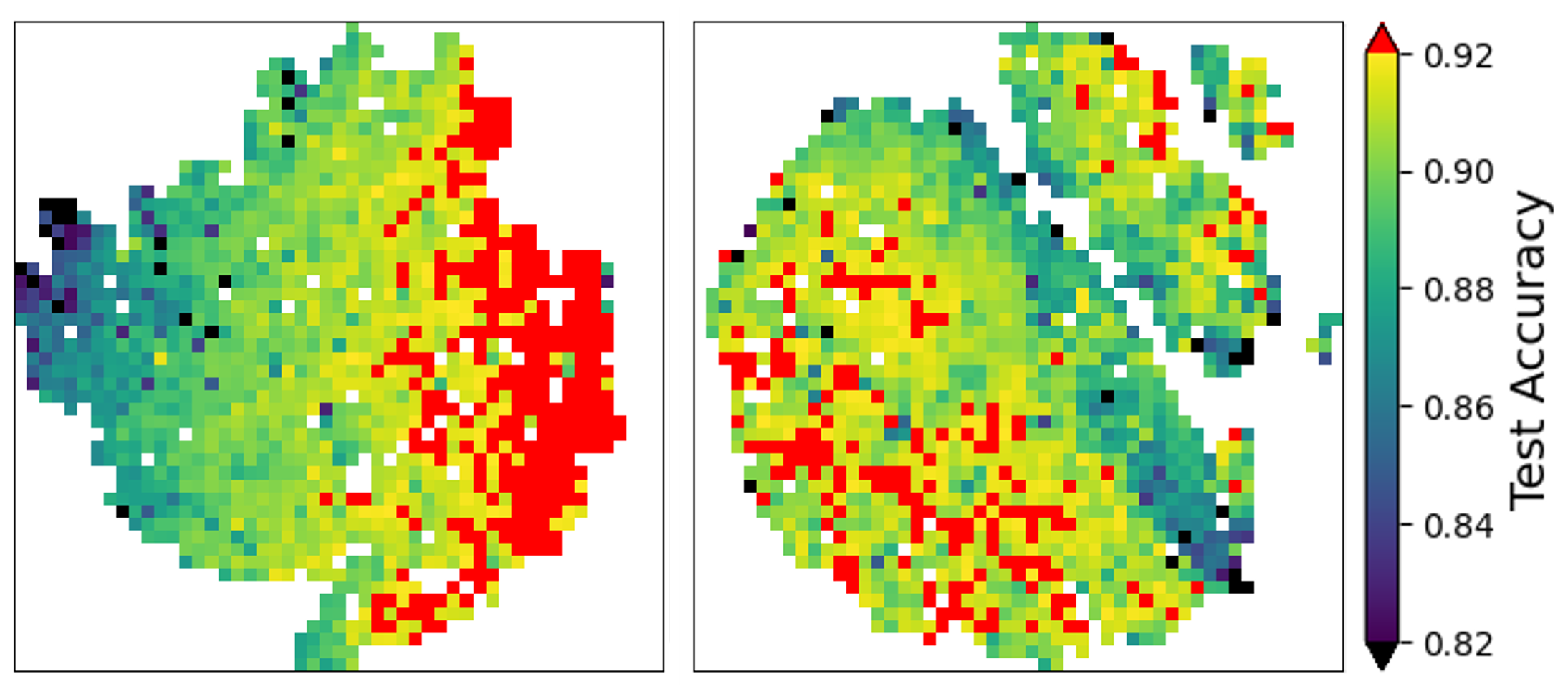}
    \caption{Visualization results of architecture representation from the pre-trained CAP (left) and baseline predictor (right). Both predictors are trained with 100 annotated architectures. We randomly sample 20,000 architectures to display their average test accuracy in each small area. Different colors denote architectures with different test accuracy.}
    \label{fig:Figure 4}
\end{figure}

\subsection{Visualization of Architecture Representation}
To demonstrate the effectiveness of the proposed context-aware self-supervised task directly, we visualize the representations of architectures generated by the pre-trained CAP and the baseline predictor using the t-SNE tool~\cite{van2008visualizing} in Figure~\ref{fig:Figure 4}. For the pre-trained CAP, architectures with close performance are embedded into nearby representations and the predictor could therefore rank architectures precisely using only a few annotated architectures. Instead, the baseline predictor fails to group architectures with similar performance well. This discrepancy is consistent with the results in the last section where the pre-trained CAP outperforms the baseline predictor. It also reflects that the proposed context-aware self-supervised task has positive impacts on enhancing the neural predictor.

\section{Conclusion} 
In this paper, we propose CAP with a context-aware self-supervised task, which leverages the rich contextual information among unlabeled architectures. The pre-trained predictor can produce expressive and generalizable representations of architectures, thus requiring fewer annotated samples. Through extensive experiments conducted on NAS-Bench-101, NAS-Bench-201 and DARTS search spaces, the proposed CAP outperforms peer neural predictors in terms of both relative ranking and searching for well-performing architectures. The ablation studies further verify the effectiveness of the context-aware self-supervised task. CAP offers a promising direction for the predictor-based NAS methods. In the future, we will explore more self-supervised tasks for neural predictors on search spaces composed of larger units instead of cell-based ones. 

\section*{Acknowledgments}
This work was supported by National Natural Science Foundation of China under Grant 62276175.

\bibliographystyle{named}
\bibliography{ijcai24}

\begin{thebibliography}{}

\bibitem[\protect\citeauthoryear{Bergstra and Bengio}{2012}]{bergstra2012random}
James Bergstra and Yoshua Bengio.
\newblock Random search for hyper-parameter optimization.
\newblock {\em Journal of machine learning research}, 13(2), 2012.

\bibitem[\protect\citeauthoryear{Chen \bgroup \em et al.\egroup }{2021a}]{chen2021contrastive}
Yaofo Chen, Yong Guo, Qi~Chen, Minli Li, Wei Zeng, Yaowei Wang, and Mingkui Tan.
\newblock Contrastive neural architecture search with neural architecture comparators.
\newblock In {\em Proc. of CVPR}, 2021.

\bibitem[\protect\citeauthoryear{Chen \bgroup \em et al.\egroup }{2021b}]{chen2021not}
Ziye Chen, Yibing Zhan, Baosheng Yu, Mingming Gong, and Bo~Du.
\newblock Not all operations contribute equally: Hierarchical operation-adaptive predictor for neural architecture search.
\newblock In {\em Proc. of ICCV}, 2021.

\bibitem[\protect\citeauthoryear{Chrabaszcz \bgroup \em et al.\egroup }{2017}]{chrabaszcz2017downsampled}
Patryk Chrabaszcz, Ilya Loshchilov, and Frank Hutter.
\newblock A downsampled variant of imagenet as an alternative to the cifar datasets.
\newblock {\em arXiv preprint arXiv:1707.08819}, 2017.

\bibitem[\protect\citeauthoryear{Chu \bgroup \em et al.\egroup }{2021}]{chu2021fairnas}
Xiangxiang Chu, Bo~Zhang, and Ruijun Xu.
\newblock Fairnas: Rethinking evaluation fairness of weight sharing neural architecture search.
\newblock In {\em Proc. of ICCV}, 2021.

\bibitem[\protect\citeauthoryear{Doersch \bgroup \em et al.\egroup }{2015}]{doersch2015unsupervised}
Carl Doersch, Abhinav Gupta, and Alexei~A Efros.
\newblock Unsupervised visual representation learning by context prediction.
\newblock In {\em Proc. of ICCV}, 2015.

\bibitem[\protect\citeauthoryear{Dong and Yang}{2019a}]{dong2019bench}
Xuanyi Dong and Yi~Yang.
\newblock Nas-bench-201: Extending the scope of reproducible neural architecture search.
\newblock In {\em Proc. of ICLR}, 2019.

\bibitem[\protect\citeauthoryear{Dong and Yang}{2019b}]{dong2019searching}
Xuanyi Dong and Yi~Yang.
\newblock Searching for a robust neural architecture in four gpu hours.
\newblock In {\em Proc. of CVPR}, 2019.

\bibitem[\protect\citeauthoryear{Eldele \bgroup \em et al.\egroup }{2021}]{eldele2021time}
Emadeldeen Eldele, Mohamed Ragab, Zhenghua Chen, Min Wu, Chee~Keong Kwoh, Xiaoli Li, and Cuntai Guan.
\newblock Time-series representation learning via temporal and contextual contrasting.
\newblock In {\em Proc. of IJCAI}, 2021.

\bibitem[\protect\citeauthoryear{Elsken \bgroup \em et al.\egroup }{2019}]{elsken2019neural}
Thomas Elsken, Jan~Hendrik Metzen, and Frank Hutter.
\newblock Neural architecture search: A survey.
\newblock {\em The Journal of Machine Learning Research}, 20(1):1997--2017, 2019.

\bibitem[\protect\citeauthoryear{Ghiasi \bgroup \em et al.\egroup }{2019}]{ghiasi2019fpn}
Golnaz Ghiasi, Tsung-Yi Lin, and Quoc~V Le.
\newblock Nas-fpn: Learning scalable feature pyramid architecture for object detection.
\newblock In {\em Proc. of CVPR}, 2019.

\bibitem[\protect\citeauthoryear{He \bgroup \em et al.\egroup }{2016}]{he2016deep}
Kaiming He, Xiangyu Zhang, Shaoqing Ren, and Jian Sun.
\newblock Deep residual learning for image recognition.
\newblock In {\em Proc. of CVPR}, 2016.

\bibitem[\protect\citeauthoryear{Jing \bgroup \em et al.\egroup }{2022}]{jing2022graph}
Kun Jing, Jungang Xu, and Pengfei Li.
\newblock Graph masked autoencoder enhanced predictor for neural architecture search.
\newblock In {\em Proc. of IJCAI}, 2022.

\bibitem[\protect\citeauthoryear{Krizhevsky}{2009}]{Krizhevsky2009LearningML}
Alex Krizhevsky.
\newblock Learning multiple layers of features from tiny images.
\newblock 2009.

\bibitem[\protect\citeauthoryear{Liu \bgroup \em et al.\egroup }{2018}]{liu2018darts}
Hanxiao Liu, Karen Simonyan, and Yiming Yang.
\newblock Darts: Differentiable architecture search.
\newblock In {\em Proc. of ICLR}, 2018.

\bibitem[\protect\citeauthoryear{Liu \bgroup \em et al.\egroup }{2019}]{liu2019auto}
Chenxi Liu, Liang-Chieh Chen, Florian Schroff, Hartwig Adam, Wei Hua, Alan~L Yuille, and Li~Fei-Fei.
\newblock Auto-deeplab: Hierarchical neural architecture search for semantic image segmentation.
\newblock In {\em Proc. of CVPR}, 2019.

\bibitem[\protect\citeauthoryear{Liu \bgroup \em et al.\egroup }{2021a}]{liu2021self}
Xiao Liu, Fanjin Zhang, Zhenyu Hou, Li~Mian, Zhaoyu Wang, Jing Zhang, and Jie Tang.
\newblock Self-supervised learning: Generative or contrastive.
\newblock {\em IEEE transactions on knowledge and data engineering}, 35(1):857--876, 2021.

\bibitem[\protect\citeauthoryear{Liu \bgroup \em et al.\egroup }{2021b}]{liu2021homogeneous}
Yuqiao Liu, Yehui Tang, and Yanan Sun.
\newblock Homogeneous architecture augmentation for neural predictor.
\newblock In {\em Proc. of ICCV}, 2021.

\bibitem[\protect\citeauthoryear{Lu \bgroup \em et al.\egroup }{2021}]{lu2021tnasp}
Shun Lu, Jixiang Li, Jianchao Tan, Sen Yang, and Ji~Liu.
\newblock Tnasp: A transformer-based nas predictor with a self-evolution framework.
\newblock {\em Proc. of NeurIPS}, 2021.

\bibitem[\protect\citeauthoryear{Luo \bgroup \em et al.\egroup }{2018}]{luo2018neural}
Renqian Luo, Fei Tian, Tao Qin, Enhong Chen, and Tie-Yan Liu.
\newblock Neural architecture optimization.
\newblock {\em Proc. of NeurIPS}, 2018.

\bibitem[\protect\citeauthoryear{Luo \bgroup \em et al.\egroup }{2020}]{luo2020semi}
Renqian Luo, Xu~Tan, Rui Wang, Tao Qin, Enhong Chen, and Tie-Yan Liu.
\newblock Semi-supervised neural architecture search.
\newblock {\em Proc. of NeurIPS}, 2020.

\bibitem[\protect\citeauthoryear{Mellor \bgroup \em et al.\egroup }{2021}]{mellor2021neural}
Joe Mellor, Jack Turner, Amos Storkey, and Elliot~J Crowley.
\newblock Neural architecture search without training.
\newblock In {\em Proc. of ICML}, 2021.

\bibitem[\protect\citeauthoryear{Mikolov \bgroup \em et al.\egroup }{2013}]{mikolov2013distributed}
Tomas Mikolov, Ilya Sutskever, Kai Chen, Greg~S Corrado, and Jeff Dean.
\newblock Distributed representations of words and phrases and their compositionality.
\newblock {\em Proc. of NeurIPS}, 2013.

\bibitem[\protect\citeauthoryear{Ning \bgroup \em et al.\egroup }{2020}]{ning2020generic}
Xuefei Ning, Yin Zheng, Tianchen Zhao, Yu~Wang, and Huazhong Yang.
\newblock A generic graph-based neural architecture encoding scheme for predictor-based nas.
\newblock In {\em Proc. of ECCV}, 2020.

\bibitem[\protect\citeauthoryear{Pham \bgroup \em et al.\egroup }{2018}]{pham2018efficient}
Hieu Pham, Melody Guan, Barret Zoph, Quoc Le, and Jeff Dean.
\newblock Efficient neural architecture search via parameters sharing.
\newblock In {\em Proc. of ICML}, 2018.

\bibitem[\protect\citeauthoryear{Real \bgroup \em et al.\egroup }{2019}]{real2019regularized}
Esteban Real, Alok Aggarwal, Yanping Huang, and Quoc~V Le.
\newblock Regularized evolution for image classifier architecture search.
\newblock In {\em Proc. of AAAI}, 2019.

\bibitem[\protect\citeauthoryear{Rendle \bgroup \em et al.\egroup }{2012}]{rendle2012bpr}
Steffen Rendle, Christoph Freudenthaler, Zeno Gantner, and Lars Schmidt-Thieme.
\newblock Bpr: Bayesian personalized ranking from implicit feedback.
\newblock {\em arXiv preprint arXiv:1205.2618}, 2012.

\bibitem[\protect\citeauthoryear{Ru \bgroup \em et al.\egroup }{2020}]{ru2020interpretable}
Binxin Ru, Xingchen Wan, Xiaowen Dong, and Michael Osborne.
\newblock Interpretable neural architecture search via bayesian optimisation with weisfeiler-lehman kernels.
\newblock In {\em Proc. of ICLR}, 2020.

\bibitem[\protect\citeauthoryear{Ru \bgroup \em et al.\egroup }{2021}]{ru2021speedy}
Robin Ru, Clare Lyle, Lisa Schut, Miroslav Fil, Mark van~der Wilk, and Yarin Gal.
\newblock Speedy performance estimation for neural architecture search.
\newblock {\em Proc. of NeurIPS}, 2021.

\bibitem[\protect\citeauthoryear{Shi \bgroup \em et al.\egroup }{2020}]{shi2020bridging}
Han Shi, Renjie Pi, Hang Xu, Zhenguo Li, James Kwok, and Tong Zhang.
\newblock Bridging the gap between sample-based and one-shot neural architecture search with bonas.
\newblock {\em Proc. of NeurIPS}, 2020.

\bibitem[\protect\citeauthoryear{Sun \bgroup \em et al.\egroup }{2019}]{sun2019evolving}
Yanan Sun, Bing Xue, Mengjie Zhang, and Gary~G Yen.
\newblock Evolving deep convolutional neural networks for image classification.
\newblock {\em IEEE Transactions on Evolutionary Computation}, 24(2):394--407, 2019.

\bibitem[\protect\citeauthoryear{Tanaka \bgroup \em et al.\egroup }{2020}]{tanaka2020pruning}
Hidenori Tanaka, Daniel Kunin, Daniel~L Yamins, and Surya Ganguli.
\newblock Pruning neural networks without any data by iteratively conserving synaptic flow.
\newblock {\em Proc. of NeurIPS}, 2020.

\bibitem[\protect\citeauthoryear{Van~der Maaten and Hinton}{2008}]{van2008visualizing}
Laurens Van~der Maaten and Geoffrey Hinton.
\newblock Visualizing data using t-sne.
\newblock {\em Journal of machine learning research}, 9(11), 2008.

\bibitem[\protect\citeauthoryear{Wei \bgroup \em et al.\egroup }{2022}]{wei2022npenas}
Chen Wei, Chuang Niu, Yiping Tang, Yue Wang, Haihong Hu, and Jimin Liang.
\newblock Npenas: Neural predictor guided evolution for neural architecture search.
\newblock {\em IEEE Transactions on Neural Networks and Learning Systems}, 2022.

\bibitem[\protect\citeauthoryear{Wen \bgroup \em et al.\egroup }{2020}]{wen2020neural}
Wei Wen, Hanxiao Liu, Yiran Chen, Hai Li, Gabriel Bender, and Pieter-Jan Kindermans.
\newblock Neural predictor for neural architecture search.
\newblock In {\em Proc. of ECCV}, 2020.

\bibitem[\protect\citeauthoryear{White \bgroup \em et al.\egroup }{2021}]{white2021bananas}
Colin White, Willie Neiswanger, and Yash Savani.
\newblock Bananas: Bayesian optimization with neural architectures for neural architecture search.
\newblock In {\em Proc. of AAAI}, 2021.

\bibitem[\protect\citeauthoryear{Williams}{1992}]{williams1992simple}
Ronald~J Williams.
\newblock Simple statistical gradient-following algorithms for connectionist reinforcement learning.
\newblock {\em Machine learning}, 8:229--256, 1992.

\bibitem[\protect\citeauthoryear{Xie \bgroup \em et al.\egroup }{2022}]{xie2022self}
Yaochen Xie, Zhao Xu, Jingtun Zhang, Zhengyang Wang, and Shuiwang Ji.
\newblock Self-supervised learning of graph neural networks: A unified review.
\newblock {\em IEEE transactions on pattern analysis and machine intelligence}, 45(2):2412--2429, 2022.

\bibitem[\protect\citeauthoryear{Xie \bgroup \em et al.\egroup }{2023}]{xie2023efficient}
Xiangning Xie, Xiaotian Song, Zeqiong Lv, Gary~G Yen, Weiping Ding, and Yanan Sun.
\newblock Efficient evaluation methods for neural architecture search: A survey.
\newblock {\em arXiv preprint arXiv:2301.05919}, 2023.

\bibitem[\protect\citeauthoryear{Xu \bgroup \em et al.\egroup }{2018}]{xu2018powerful}
Keyulu Xu, Weihua Hu, Jure Leskovec, and Stefanie Jegelka.
\newblock How powerful are graph neural networks?
\newblock In {\em Proc. of ICLR}, 2018.

\bibitem[\protect\citeauthoryear{Xu \bgroup \em et al.\egroup }{2021}]{xu2021renas}
Yixing Xu, Yunhe Wang, Kai Han, Yehui Tang, Shangling Jui, Chunjing Xu, and Chang Xu.
\newblock Renas: Relativistic evaluation of neural architecture search.
\newblock In {\em Proc. of CVPR}, 2021.

\bibitem[\protect\citeauthoryear{Yan \bgroup \em et al.\egroup }{2020}]{yan2020does}
Shen Yan, Yu~Zheng, Wei Ao, Xiao Zeng, and Mi~Zhang.
\newblock Does unsupervised architecture representation learning help neural architecture search?
\newblock {\em Proc. of NeurIPS}, 2020.

\bibitem[\protect\citeauthoryear{Yan \bgroup \em et al.\egroup }{2021}]{yan2021cate}
Shen Yan, Kaiqiang Song, Fei Liu, and Mi~Zhang.
\newblock Cate: Computation-aware neural architecture encoding with transformers.
\newblock In {\em Proc. of ICML}, 2021.

\bibitem[\protect\citeauthoryear{Yi \bgroup \em et al.\egroup }{2023}]{yi2023nar}
Yun Yi, Haokui Zhang, Wenze Hu, Nannan Wang, and Xiaoyu Wang.
\newblock Nar-former: Neural architecture representation learning towards holistic attributes prediction.
\newblock In {\em Proc. of CVPR}, 2023.

\bibitem[\protect\citeauthoryear{Ying \bgroup \em et al.\egroup }{2019}]{ying2019bench}
Chris Ying, Aaron Klein, Eric Christiansen, Esteban Real, Kevin Murphy, and Frank Hutter.
\newblock Nas-bench-101: Towards reproducible neural architecture search.
\newblock In {\em Proc. of ICML}, 2019.

\bibitem[\protect\citeauthoryear{Zela \bgroup \em et al.\egroup }{2020}]{zela2020surrogate}
Arber Zela, Julien Siems, Lucas Zimmer, Jovita Lukasik, Margret Keuper, and Frank Hutter.
\newblock Surrogate nas benchmarks: Going beyond the limited search spaces of tabular nas benchmarks.
\newblock {\em arXiv preprint arXiv:2008.09777}, 2020.

\bibitem[\protect\citeauthoryear{Zheng \bgroup \em et al.\egroup }{2022}]{zheng2022neural}
Xiawu Zheng, Xiang Fei, Lei Zhang, Chenglin Wu, Fei Chao, Jianzhuang Liu, Wei Zeng, Yonghong Tian, and Rongrong Ji.
\newblock Neural architecture search with representation mutual information.
\newblock In {\em Proc. of CVPR}, 2022.

\bibitem[\protect\citeauthoryear{Zoph \bgroup \em et al.\egroup }{2018}]{zoph2018learning}
Barret Zoph, Vijay Vasudevan, Jonathon Shlens, and Quoc~V Le.
\newblock Learning transferable architectures for scalable image recognition.
\newblock In {\em Proc. of CVPR}, 2018.

\end{thebibliography}

\end{document}